\documentclass{article} 
\usepackage[final]{colm2026_conference}
\usepackage{mathtools}
\usepackage{amsmath,amssymb}
\usepackage{microtype}
\usepackage{hyperref}
\usepackage{url}
\usepackage{booktabs}
\usepackage{graphicx}
\usepackage{subcaption}
\usepackage{tikz}
\usepackage{changepage}
\usepackage{algorithm}
\usepackage{algpseudocode}
\usepackage[most]{tcolorbox}

\usetikzlibrary{arrows.meta,positioning,calc}

\usepackage{lineno}

\definecolor{darkblue}{rgb}{0, 0, 0.5}
\hypersetup{colorlinks=true, citecolor=darkblue, linkcolor=darkblue, urlcolor=darkblue}

\title{MathDuels: A Self-Play Benchmark That Grows}

\author{%
  Zhiqiu Xu$^{1,3}$\thanks{Equal contribution.\\Correspondence to: Zhiqiu Xu \texttt{<oscarxzq@seas.upenn.edu>}, Shibo Jin \texttt{<shibo517@seas.upenn.edu>}.} \qquad
  Shibo Jin$^{1}$\footnotemark[1] \qquad
  Shreya Arya$^{2,3}$ \qquad
  Mayur Naik$^{1,3}$ \\
  \\
  $^{1}$Department of Computer and Information Science, University of Pennsylvania \\
  $^{2}$Department of Mathematics, University of Pennsylvania \\
  $^{3}$Rabdos AI
}

\begin{document}

\ifcolmsubmission
\linenumbers
\fi
\maketitle
\vspace{-1.0em}
{\centering
{\hypersetup{urlcolor=[RGB]{0,102,103}}%
\href{https://mathduels.ai}{\textbf{Leaderboard:} \texttt{mathduels.ai}}}\par
}
\vspace{1.0em}

\begin{abstract}
As frontier language models attain near-ceiling performance on static mathematical benchmarks, existing evaluations are increasingly unable to differentiate model capabilities, largely because they cast models solely as solvers of fixed problem sets.
We introduce MathDuels, a self-play benchmark in which models occupy dual roles: each authors math problems under adversarial prompting and solves problems authored by every other participant.
Problems are produced through a three-stage generation pipeline (meta-prompting, problem generation, and difficulty amplification), and validated by an independent verifier that excludes ill-posed questions.
A Rasch model~\citep{rasch1993probabilistic} jointly estimates solver abilities and problem difficulties; author quality is derived from the difficulties of each model's authored problems.
Experiments across 19 frontier models reveal that authoring and solving capabilities are partially decoupled,
and that dual-role evaluation reveals capability separations invisible in single-role benchmarks.
As newer models enter the arena, they produce problems that defeat previously dominant solvers, so the benchmark's difficulty co-evolves with participant strength rather than saturating at a fixed ceiling.
We host a public leaderboard that updates as new models are released.
\end{abstract}

\section{Introduction}

Static benchmarks for mathematical reasoning in LLMs are saturating faster than new problems can be produced.
Benchmarks such as MATH~\citep{hendrycks2021measuring} and GSM8K~\citep{cobbe2021training} no longer provide the headroom they once did for separating frontier systems.
The community therefore turned to live competition problems, but even these are becoming less durable: recent competition-based evaluation results report strong performance on newly released sets, including AIME~2026~\citep{balunovic2025matharena}.
Model capabilities now advance faster than the supply of fresh, discriminative problems, and no fixed or annually replenished problem pool can sustain meaningful separation for long.
This has pushed evaluation toward newly authored, research-level problems designed to remain difficult even for strong models~\citep{glazer2024frontiermath}.
When benchmark builders must target the boundary of current model capability, constructing problems that remain discriminative becomes extraordinarily difficult.

Humans, in short, find it increasingly hard to supply problems difficult enough to differentiate the strongest models.
But this predicament is not new.
The question of how to separate two practitioners who both operate at the frontier of a discipline is one that mathematicians confronted centuries ago, and they arrived at an elegant answer: let them test each other.
In 1535, the Venetian mathematician Niccol\`{o} Tartaglia received a challenge from Antonio Maria Fior: each would deposit thirty problems with a notary, and whoever solved more of the other's set within fifty days would win~\citep{toscano2020secret}.
Victory required not just solving skill but the ability to \emph{construct} problems that exceeded the opponent's reach.
Tartaglia solved all thirty of Fior's problems in a short period; Fior could not solve a single one of Tartaglia's.
By forcing each to probe the boundary of the other's knowledge, the duel exposed a gap that no static test could have surfaced.

We bring this principle to LLM evaluation.
We introduce \textbf{MathDuels}, a self-play benchmark in which models occupy both roles simultaneously.
Each model authors math problems under a fixed budget and solves problems authored by every other model.
A model is rewarded both for solving opponents' problems and for authoring problems that resist solution.
As stronger models enter the arena, they generate problems that defeat previously dominant solvers, so the benchmark's difficulty co-evolves with participant strength rather than saturating at a fixed ceiling.

Concretely, a single evaluation round proceeds as follows.
Each of $N$ participating models authors $K$ problems via a prompted generation pipeline that includes adversarial difficulty amplification.
Every non-author model then attempts each problem.
Answers are judged automatically through symbolic verification, and problems that defeat any solver undergo a validity check to filter ill-posed or ambiguous questions.
The resulting outcome matrix feeds a Rasch model that jointly estimates solver abilities and problem difficulties, with author quality derived from the difficulties of each model's authored problems.

\begin{figure*}[t]
    \centering
    \includegraphics[width=1.00\textwidth]{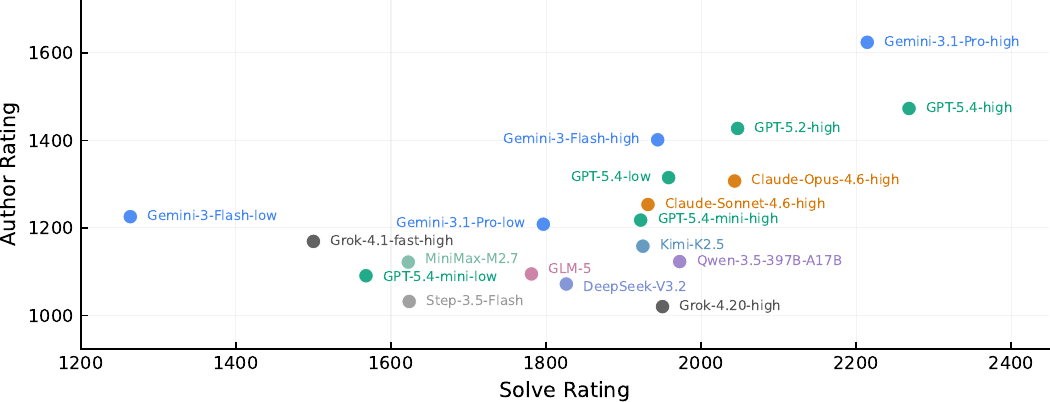}
    \caption{Solver rating vs.\ Author rating for 19 frontier models.}\label{fig:scatter}
    \end{figure*}

Our experiments across 19 frontier models reveal two findings that motivate the self-play design.
First, solving capability and authoring capability are partially decoupled: strong solvers are not necessarily strong authors, suggesting these are partially independent axes of mathematical competence that single-role benchmarks conflate or ignore entirely.
Second, as stronger models enter the arena, they generate problems that defeat previously dominant solvers, so the benchmark's difficulty co-evolves with participant strength rather than saturating at a fixed ceiling.
\paragraph{Contributions.} We summarize the contributions of our work:
\begin{itemize}
    \item We propose a self-play benchmark protocol for mathematical LLM evaluation in which models are scored in dual roles, as both problem authors and solvers, with automated problem validity filtering.
    \item We conduct experiments across 19 frontier models and show that dual-role evaluation reveals capability separations invisible in single-role benchmarks, that authoring and solving abilities are partially decoupled, and that the arena resists saturation by design.
    \item We host a public leaderboard which will be updated when new models are released.
\end{itemize}

\section{The MathDuels protocol}
\label{sec:protocol}

We formalize an analogous protocol for language models.
A MathDuels run takes a set of $N$ models and subjects each to evaluation in two roles: as an \emph{author} of math problems and as a \emph{solver} of problems authored by others.
The protocol proceeds in four phases (generation, solving, judging, and ranking), each designed to be fully automated and reproducible without human annotation in the loop.
Algorithm~\ref{alg:protocol} gives a high-level pseudocode summary; the subsections that follow expand each phase.

\algrenewcommand{\algorithmiccomment}[1]{\hfill\textit{$\triangleright$~#1}}
\begin{algorithm}[t]
\caption{The MathDuels protocol}\label{alg:protocol}
\small
\begin{algorithmic}[1]
\Statex \textbf{Input:} Set of models $\mathcal{M}=\{m_1,\dots,m_N\}$; problems per model $K$; domains $\{\delta_k\}$
\Statex \textbf{Output:} Composite ratings $\{R_m\}_{m=1}^{N}$
\State \textbf{for all} $m \in \mathcal{M},\; k = 1,\dots,K$: \hfill$\triangleright$\makebox[6.5cm][l]{\;\textit{Phase 1: Generation}}
\State \hspace{\algorithmicindent} $p \coloneqq (q, g) \gets \Call{Generate}{m, \delta_k}$
\Statex \vspace{-0.5\baselineskip}
\State \textbf{for all} $(m, p)$ with $m \neq \mathrm{author}(p)$: \hfill$\triangleright$\makebox[6.5cm][l]{\;\textit{Phase 2: Solving}}
\State \hspace{\algorithmicindent} $y_{mp} \gets \mathbf{1}[\Call{Solve}{m, q_p} \equiv g_p]$
\Statex \vspace{-0.5\baselineskip}
\State \textbf{for all} $p$ where $\exists\, m : y_{mp} = 0$: \hfill$\triangleright$\makebox[6.5cm][l]{\;\textit{Phase 3: Verification}}
\State \hspace{\algorithmicindent} $(v, a^*) \gets \Call{Verify}{q_p, g_p, \{(a_i, r_i)\}}$
\State \hspace{\algorithmicindent} \textbf{if} $v{=}0$ drop $p$, \textbf{else} $g_p \gets a^*$
\Statex \vspace{-0.5\baselineskip}
\State Fit $\{s_m\}, \{d_p\}$ via \hfill$\triangleright$\makebox[6.5cm][l]{\;\textit{Phase 4: Ranking}}
\Statex \hspace{\algorithmicindent} $\max \sum_{(m,p)} \bigl[y_{mp}\log\sigma(s_m {-} d_p) + (1{-}y_{mp})\log\sigma(d_p {-} s_m)\bigr] - \lambda \!\sum_p d_p^2$
\State $R_m^{\mathrm{solve}} \gets R_{\mathrm{anchor}} + C_{\mathrm{elo}}(s_m - s_{\mathrm{anchor}})$;\quad $R_m^{\mathrm{author}} \gets \mathrm{mean}\!\big(\{C_{\mathrm{elo}}\,\tilde{d}_p : \mathrm{author}(p)=m\}\big)$
\State $R_m \gets w_{\mathrm{solve}}\,R_m^{\mathrm{solve}} + w_{\mathrm{author}}\,R_m^{\mathrm{author}}$
\end{algorithmic}
\end{algorithm}


\subsection{Arena setup}
\label{sec:setup}

Let $\mathcal{M}=\{m_1,\dots,m_N\}$ be the set of participating models and let $K$ be the number of problems each model generates, giving $NK$ problems in total. Every model attempts every problem it did not author, so for each solver--problem pair $(m,p)$ with $m\neq\mathrm{author}(p)$ we observe a binary outcome
\[
y_{mp}\in\{0,1\},\qquad y_{mp}=1\;\Longleftrightarrow\;\text{model }m\text{ solves problem }p\text{ correctly}.
\]
These outcomes populate a solve matrix $Y\in\{0,1\}^{N\times NK}$ whose entries are defined only when $m\neq\mathrm{author}(p)$. The full evaluation pipeline is:
\[
\mathcal{M}
\;\xrightarrow{\text{generate}}\;
NK\text{ problems}
\;\xrightarrow{\text{attempt}}\;
Y
\;\xrightarrow{\text{rank}}\;
\{R_m\}_{m=1}^{N},
\]
where $R_m$ is the final composite rating of model $m$. Sections~\ref{sec:generation}--\ref{sec:ranking} describe each stage.

\subsection{Problem generation}
\label{sec:generation}

Generating problems that are simultaneously hard, well-posed, and unambiguously verifiable is the central design challenge.
We use a three-stage pipeline: meta-prompting, problem generation, and difficulty amplification.
The guiding design principle is minimal human steering: beyond specifying that problems should be difficult and framing the task as a competitive arena, the prompts impose no constraints on mathematical content, style, or strategy, letting each model explore its own notion of what makes a problem hard.

\paragraph{Stage~1: meta-prompting.} We first ask the model to write a prompt for authoring a problem, given a mathematical domain $\delta_k$. This forces the model to reason about what makes a problem hard before constructing one, and because each sampled meta-prompt encodes a different strategy, the approach naturally boosts diversity across the generated problem set and lets the model plan for domain-specific variation.

\paragraph{Stage~2: problem generation.} The prompt generated in Stage~1 is fed back to the same model. The model's output contains a problem statement $q_k$ and a gold answer expressed as a single scalar mathematical expression $g_k$.

\paragraph{Stage~3: difficulty amplification.} Each problem may undergo one or more hardening rounds in which the authoring model produces a variant that elicits deeper reasoning.

\subsection{Solving}
\label{sec:solving}

For each problem, every non-author model attempts a solution. Each solver returns a final answer $a$ and a reasoning trace $r$ (the trace is used later in verification when answers disagree or all solvers fail; see Section~\ref{sec:judging}).

Correctness is determined by symbolic equivalence checking against the problem's gold answer $g$, yielding a binary outcome $y_{mp}\in\{0,1\}$ for each solver--problem pair.
These outcomes populate the solve matrix used in ranking (Section~\ref{sec:ranking}).

\subsection{Verification}
\label{sec:judging}

\begin{adjustwidth}{.5cm}{.5cm}
\emph{''in case of disagreement a public debate should be held''~\citep[p.~13]{toscano2020secret}.}
\end{adjustwidth}

When at least one solver answer is incorrect, we run a secondary verification step. Let $\mathcal{A}_p = \{(a_i, r_i)\}$ denote the set of unique solver answers and their reasoning traces for problem $p$. Verification is triggered only when disagreement exists:
\[
\exists\; i : y_i = 0 \;\;\Longrightarrow\;\; (v,\, a^*) \;=\; \texttt{Verify}\!\left(q_p,\; g_p,\; \mathcal{A}_p\right), \qquad v\in\{0,1\},\quad a^*\in\mathcal{A}_p.
\]
The verifier receives the problem statement $q_p$, the author's gold answer $g_p$, and the full candidate set $\mathcal{A}_p$, then (i) checks whether $q_p$ has a unique well-defined solution and (ii) selects the correct answer from the candidates. Here $v=1$ means the problem is valid, and $a^*$ is the verifier-selected answer. If $v=0$, the problem is excluded from scoring; otherwise $a^*$ replaces $g_p$ as the reference answer.

\subsection{Ranking}
\label{sec:ranking}

Each solve attempt in the outcome matrix is a binary observation: model $m$ either solves problem $p$ or does not.
This is precisely the setting of item response theory (IRT), where test-takers of varying ability face items of varying difficulty.
We adopt the Rasch model~\citep{rasch1993probabilistic}, which is mathematically identical to a Bradley--Terry\citep{bradley1952rank} model applied to solver--problem pairs: each solve attempt is a ``match'' between a solver and a problem, and the probability of a correct answer depends only on the difference between the solver's ability and the problem's difficulty.

\paragraph{The model.}
Let $s_m$ denote the ability of solver $m$ and $d_p$ the difficulty of problem $p$, both on a shared logit scale.
The probability of a correct response is
\begin{equation}
\label{eq:rasch}
P(y_{mp}=1) \;=\; \sigma(s_m - d_p) \;=\; \frac{1}{1 + e^{-(s_m - d_p)}},
\end{equation}
where $\sigma(\cdot)$ is the logistic function.
When $s_m \gg d_p$ the solver almost certainly succeeds; when $d_p \gg s_m$ the problem is too hard.
The key property is that correctness on a hard problem (high $d_p$) is more informative about solver ability than correctness on an easy one, and vice versa for problem difficulty.

\paragraph{Estimation.}
All solver abilities $\{s_m\}$ and problem difficulties $\{d_p\}$ are estimated jointly by maximizing the log-likelihood over all attempts (excluding each author's own problems):
\begin{equation}
\label{eq:loglik}
\ell(\mathbf{s}, \mathbf{d}) \;=\; \sum_{(m,p)\in\mathcal{O}} \Big[ y_{mp}\log\sigma(s_m - d_p) + (1-y_{mp})\log(1-\sigma(s_m - d_p)) \Big] - \lambda\sum_p d_p^2,
\end{equation}
where $\mathcal{O}$ is the set of observed solver--problem pairs after excluding invalid problems and self-authored attempts, and $\lambda$ is a mild $\ell_2$ regularizer on problem difficulties that prevents divergence for problems solved by all or no solvers.
Since the Rasch likelihood is invariant to a global shift of all parameters, we anchor the scale by fixing a reference model's solver rating, making scores comparable across runs.
Logit-scale parameters are converted to ratings via $R = R_{\mathrm{anchor}} + C_{\mathrm{elo}}(s_m - s_{\mathrm{anchor}})$ where $C_{\mathrm{elo}} = 400/\!\ln 10$ is the standard Elo scaling constant.

\paragraph{Author rating.}
Author quality is not a fitted parameter but a derived summary statistic: $R_m^{\mathrm{author}} = \frac{1}{|\mathcal{P}_m|}\sum_{p \in \mathcal{P}_m} D_p$, the mean rating-scaled difficulty of the problems authored by model $m$.
Two safeguards prevent degenerate author credit.
First, problems excluded by the verifier as invalid (Section~\ref{sec:judging}) are removed from all likelihood computations, so they contribute to neither solver abilities nor problem difficulties.
Second, if the author's own submitted answer is judged incorrect, that problem's difficulty is capped at $\min(d_p,\, \bar{d}_m^{\,\mathrm{correct}})$ when counting towards author's rating, where $\bar{d}_m^{\,\mathrm{correct}}$ is the mean difficulty of the author's gold-correct problems; this prevents an author from inflating its rating with hard but incorrectly specified problems.

\paragraph{Composite rating.}
The final rating combines both roles with equal weight so that problem construction is treated as equally important as problem solving:
\[
R_m \;=\; \tfrac{1}{2} \,R_m^{\mathrm{solve}} + \tfrac{1}{2}\,R_m^{\mathrm{author}}.
\]

\paragraph{Confidence intervals.}
We report stratified bootstrap 95\% confidence intervals ($B = 10,000$ resamples, stratified by author to preserve per-author problem budgets).
For a given rating $\theta_m$, the interval is \[\mathrm{CI}_{95}(\theta_m) = [\,\hat{\theta}_m^{*(0.025)},\;\hat{\theta}_m^{*(0.975)}\,]\] where $\hat{\theta}_m^{*(\alpha)}$ denotes the $\alpha$-quantile of the bootstrap distribution.
\label{eq:bootstrap_ci}

\section{Experiments}
\label{sec:experiments}

We evaluate MathDuels on a diverse set of frontier language models.
The goals are to test whether the protocol produces a discriminative ranking, to examine the relationship between authoring and solving, and to quantify how the composite rating reshapes the leaderboard relative to accuracy alone.

\subsection{Setup}
\label{sec:setup_exp}

\paragraph{Participants.}
We evaluate 19 frontier models from nine providers~\citep{openai2025gpt52,openai2026gpt54,openai2026gpt54mini,google2026gemini31pro,google2025gemini3flash,anthropic2026claudeopus46,anthropic2026claudesonnet46,xai2025grok41fast,xai2026grok420,alibaba2026qwen35,moonshot2026kimik25,deepseek2025deepseekv32,zhipu2026glm5,minimax2026m27,stepfun2026step35flash}. Several families are tested at both high and low reasoning effort, allowing us to probe how thinking time affects authoring and solving independently.

\paragraph{Configuration.}
Each model authors 30 problems per run, uniformly distributed across six broad mathematical domains -- analysis, algebra, geometry \& topology, discrete mathematics, probability \& statistics, and applied \& computational mathematics -- via domain-specific meta-prompts (Appendix~\ref{sec:appendix_taxonomy}).
Every problem is generated from the 3-step protocol introduced in Section~\ref{sec:generation}. We use \texttt{math-verify}~\citep{math_verify} for expression equivalence checking. 
We select \texttt{GPT-5.4-high} as backbone of verifier, which is ablated in Section~\ref{sec:ablation_verifier}.
All inference calls use temperature 1.0.
\paragraph{Scale.}
The 19 models produce 570 problems in total.
Of these, 11 are excluded by the verification layer.
The remaining 559 valid problems yield 10{,}062 binary solve observations (each model attempts roughly 530 problems authored by others), forming the basis of all reported statistics.

\subsection{Results}
\label{sec:results}

\begin{figure*}[t]
\centering
\includegraphics[width=0.98\textwidth]{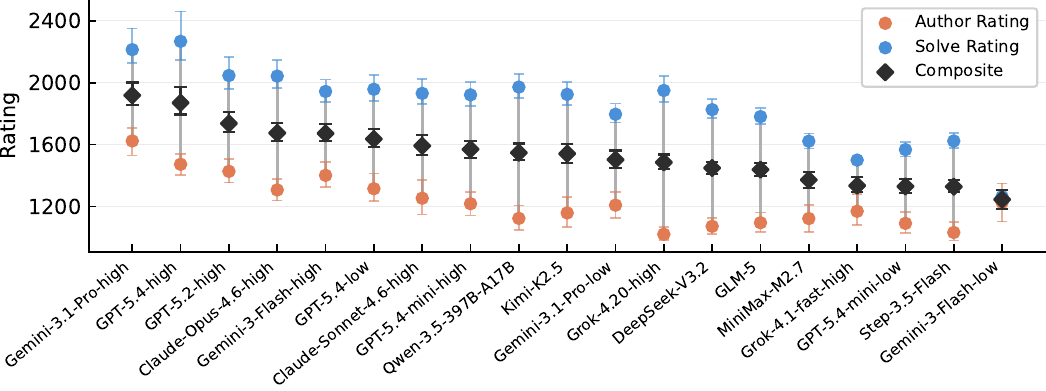}
\caption{The 19 frontier models sorted by composite rating. Whiskers show bootstrapped 95\% confidence intervals (10,000 iterations, stratified by author).}\label{fig:elo_dumbbell}
\end{figure*}

\paragraph{Composite ratings and solver--author decoupling.}
The Rasch model (Section~\ref{sec:ranking}) jointly estimates solver abilities and problem difficulties, with \texttt{Grok-4.1-fast-high} anchored at 1500.
Figure~\ref{fig:elo_dumbbell} shows the composite rating and its bootstrapped 95\% confidence intervals.
Interestingly, the best solver is not the best overall model.
\texttt{GPT-5.4-high} leads all participants in solver rating, yet \texttt{Gemini-3.1-Pro-high} takes the top composite rank because it is by far the strongest author in the field: its authored problems have the lowest average solve rate of any participant (62.9\%).
This authoring advantage is invisible to any solve-only benchmark.
Notably, \texttt{Grok-4.20-high} has the largest gap in solver and author rating. We analyzed its authored problems along with \texttt{Gemini-3.1-Pro-high}'s in Section~\ref{sec:case_study}.


\begin{figure}[t]
    \centering
    \begin{minipage}[b]{0.63\linewidth}
      \centering
      \includegraphics[width=\linewidth]{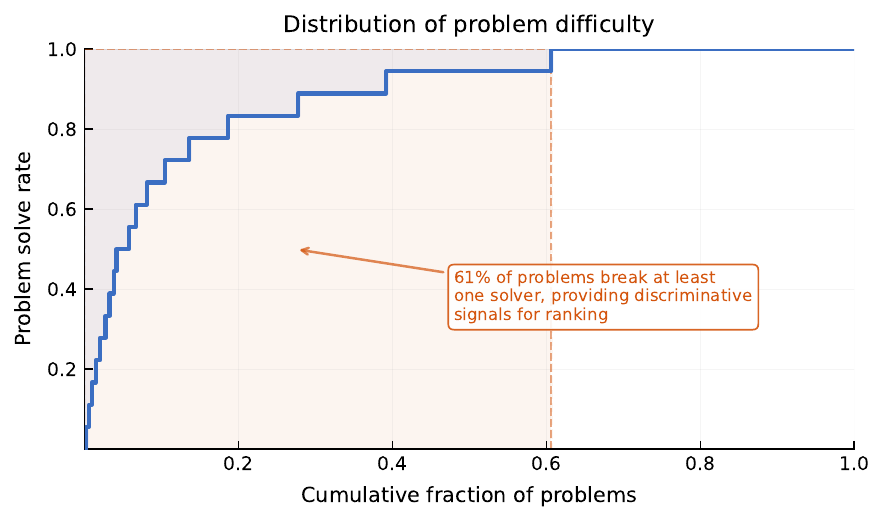}
      \subcaption{Problem solve-rate CDF.}\label{fig:difficulty_cdf}
    \end{minipage}\hfill
    \begin{minipage}[b]{0.35\linewidth}
      \centering
      \includegraphics[width=\linewidth]{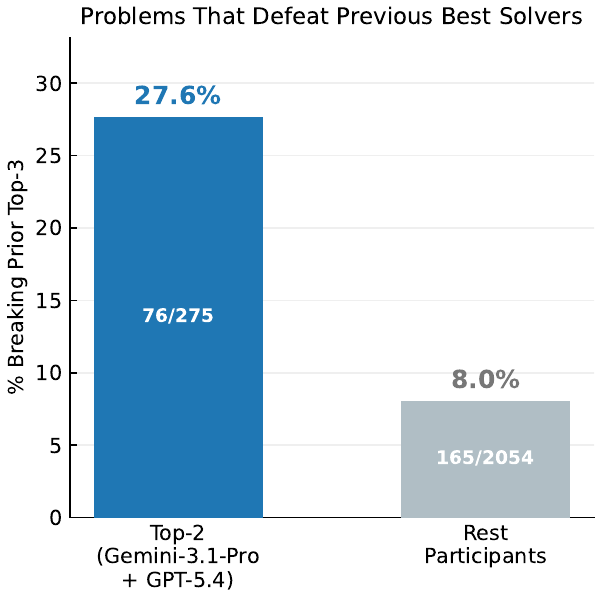}
      \subcaption{Breaking the prior top-3.}\label{fig:newcomers}
    \end{minipage}
    \caption{(a) Cumulative distribution of problem difficulty. 39\% of problems are solved by all solvers. Authoring hard problems that break other models is more challenging than solving problems. (b) Problems authored by the two strongest newcomers break prior top-3 solvers at 3.4$\times$ the rate of problems from the rest of the participants.}\label{fig:difficulty}
    \end{figure}

\paragraph{Problem difficulty distribution.}
Figure~\ref{fig:difficulty}(a) plots the cumulative distribution of problem solve rates across the pool.
Despite the three-prompt generation pipeline, roughly 39\% of valid problems are solved by every non-author solver, offering no discriminative signal on their own.
Yet the remaining problems do carry signal: the Rasch model extracts reliable ranking differences from the discriminating tail below full solve rate, separating both solvers and authors.
That tail is disproportionately shaped by a few strong authors (\texttt{Gemini-3.1-Pro-high} contributes the widest spread and highest median difficulty, consistent with its dominant Author rating), and, as we show next, it deepens each time a stronger model enters the arena.

Figure~\ref{fig:difficulty}(b) illustrates this directly.
We define the ``prior top-3'' solvers as the three highest-accuracy models excluding the two newest entrants (\texttt{Gemini-3.1-Pro-high} and \texttt{GPT-5.4-high}), yielding \texttt{GPT-5.2-high}, \texttt{Claude-Opus-4.6-high}, and \texttt{Qwen-3.5-397B-A17B}.
Among the 275 problems authored by the two newcomers, 27.6\% defeat at least one prior top-3 solver -- more than three times the 8.0\% rate for problems authored by the remaining 17 participants.
The implication is practical: when a stronger model joins the arena, it does not merely solve more problems; it also \emph{authors} harder ones that expose weaknesses in the previous best, automatically raising the difficulty ceiling without any manual curation.

\subsection{Problem analysis}
\label{sec:case_study}

\noindent Solve rates vary substantially across the six mathematical domains in our taxonomy (Table~\ref{tab:problem-taxonomy}): Discrete Mathematics is the hardest (82.5\% average accuracy) and Probability \& Statistics the easiest (92.2\%).
This reflects not an intrinsic difficulty ranking of fields, but rather the kind of problems the pipeline tends to produce in each.
Hard discrete problems are typically short and easy to verify, yet difficult because their structure is not explicit in the problem statement; the solver must discover the right counting framework before any standard method applies.
Problems in algebraic topology or category theory, by contrast, are harder to \emph{author} correctly but, once the intended machinery is recognized, often direct to solve.

For example, \texttt{GPT-5.2-high} authored a topology problem on the invariant-factor decomposition of $\pi_2$ of a mapping torus built from $\bigvee_{i=1}^4 S_i^2$: heavy machinery, 17/19 solve rate.
The same model posed a combinatorics question asking for the number of $3$-colorings of a regular $12$-gon under dihedral symmetry with exact color-count and adjacency constraints: two sentences to state, 11/19 solve rate.
The topology problem advertises its difficulty; the combinatorics problem hides it (see Appendix~\ref{sec:appendix_taxonomy} for further analysis).

At the level of individual problems, strong authors impose structural constraints that force solvers to discover the right framework before any calculation can begin, while weak authors default to textbook templates that test execution alone. We provide analysis on more generated problems and authoring patterns in Appendix~\ref{sec:appendix_authoring_styles}.

\section{Ablations}
\label{sec:analysis}

\subsection{Generation pipeline}
\label{sec:ablation_generation}

The full generation pipeline has three components: meta-prompting, problem generation, and difficulty amplification.
To isolate the contribution of each, we run the four strongest authors (\texttt{Gemini-3.1-Pro-high}, \texttt{Gemini-3-Flash-high}, \texttt{GPT-5.4-high}, and \texttt{GPT-5.2-high}) under three pipeline configurations: (1)~\emph{direct generation}, where the model receives a domain specification and produces a problem in a single step with no meta-prompting or evolution; (2)~\emph{meta-prompted generation}, which adds the two-stage meta-prompting pipeline; and (3)~the \emph{full pipeline} with meta-prompting and difficulty amplification.

\begin{table}[h]
\centering
\small
\begin{tabular}{@{}lcccc@{}}
\toprule
Pipeline & \multicolumn{2}{c}{w/o code execution} & \multicolumn{2}{c}{w/ code execution} \\
\cmidrule(lr){2-3} \cmidrule(lr){4-5}
 & Errors / 432 & Error rate & Errors / 432 & Error rate \\
\midrule
1-step (direct generation) & 22 & 5.1\% & 13 & 3.0\% \\
2-step (meta-prompting) & 41 & 9.5\% & 30 & 6.9\% \\
3-step (meta-prompting + amplification) & 108 & 25.0\% & 91 & 21.1\% \\
3-step w/ code execution & - & - & 84 & 21.8\% \\
\bottomrule
\end{tabular}
\caption{Generation pipeline ablation. Each configuration is evaluated across 432 model-pair observations. Stronger pipelines yield more errors. Code execution at solve time provides only a modest reduction in error rate.}
\label{tab:pipeline_ablation}
\end{table}

Table~\ref{tab:pipeline_ablation} reports aggregate error counts across all model-pair evaluations under each configuration, both without and with code execution enabled at either solving time or authoring time.
The result shows that each additional pipeline stage produces harder problems and roughly doubles the error rate.
Notably, because the authoring side never invokes code execution, the resulting problems predominantly test mathematical reasoning rather than large-scale computation;
tools therefore reduce the error rate by only a modest margin (e.g., from 25.0\% to 21.1\% at the hardest tier), confirming that the benchmark's discriminative power comes from reasoning difficulty, not from numerical bottlenecks that a calculator could bypass.
When code execution is activated on the author side, the error rate remains unchanged, confirming that problem difficulty is intrinsic to the underlying mathematics rather than something that can be inflated through tools during authoring.

\subsection{Ranking stability}
\label{sec:ablation_budget}

Following Chatbot Arena~\citep{chiang2024chatbot}, we derive each model's rank range from 10{,}000 stratified bootstrap iterations. Specifically, the worst-case rank range is computed by comparing each model's lower confidence bound against every other model's upper confidence bound, yielding the widest plausible interval of ranks a model could occupy under statistical uncertainty.
At $k{=}30$ across all 19 models, the mean rank range is 5.05 out of 19. We present top-5 per-model ranges in Table~\ref{tab:rank_ranges_partial} (full table in Appendix~\ref{sec:appendix_rank_ranges}). The top 2 models and bottom 4 models have non-ambiguous clusters, while mid-tier models exhibit wider, overlapping ranges.
\begin{table}[h]
\centering
\small
\begin{tabular}{@{}rlccccc@{}}
\toprule
\# & Model & Solve & Author & Composite & 95\% CI & Range \\
\midrule
1  & \texttt{Gemini-3.1-Pro-high}    & 2214 & 1624 & 1919 & [1856, 2000] & 1--2  \\
2  & \texttt{GPT-5.4-high}           & 2268 & 1473 & 1870 & [1798, 1975] & 1--3  \\
3  & \texttt{GPT-5.2-high}           & 2047 & 1427 & 1737 & [1675, 1807] & 2--6  \\
4  & \texttt{Claude-Opus-4.6-high}   & 2043 & 1307 & 1675 & [1622, 1740] & 3--8  \\
5  & \texttt{Gemini-3-Flash-high}    & 1944 & 1401 & 1673 & [1620, 1732] & 3--8  \\
\bottomrule
\end{tabular}
\caption{Top-5 model ratings with 95\% CI and worst-case ranking range.}
\label{tab:rank_ranges_partial}
\end{table}

\subsection{Verifier backbone robustness}
\label{sec:ablation_verifier}

The verification layer (Section~\ref{sec:judging}) uses a single verifier model (\texttt{GPT-5.4-high}) to filter ill-posed problems.
We test robustness by replaying the verification step with alternative verifier backbones and comparing the resulting exclusion sets and downstream rankings.

We replay the full verification pass on 414 problems using \texttt{Gemini-3.1-Pro-high} as an alternative backbone.
Table~\ref{tab:verifier_agreement} reports agreement between the two verifiers across two steps.

\begin{table}[h]
\centering
\small
\begin{tabular}{@{}lc@{}}
\toprule
Metric & Agreement \\
\midrule
Same exclusion decision & 97.5\% \\
Same selected answer & 99.4\% \\
\bottomrule
\end{tabular}
\caption{Verifier backbone agreement on 414 problems. ``Same exclusion decision'' measures whether both verifiers agree on whether a problem should be excluded. ``Same selected answer'' measures concordance on the chosen candidate among non-excluded problems, using normalized text and symbolic matching.}
\label{tab:verifier_agreement}
\end{table}

The two verifiers agree on the exclude-or-keep decision for 97.5\% of problems, and among problems both retain, they select answers whose normalized forms match 99.4\% of the time.
Two problems exhibit answer disagreement between the verifier backbones. We hand-verified both cases and found that \texttt{GPT-5.4} and \texttt{Gemini-3.1-Pro} each produced a false answer on one of the two problems. After majority sampling---drawing two additional attempts per verifier and taking the plurality vote---both models flipped to the correct answer on their respective failure case.
A structurally different verifier backbone reaches near-identical judgments confirm that verification is not an artifact of a single model's idiosyncrasies, and the downstream rankings would be essentially unchanged under either.

\section{Related work}
\label{sec:related}

\paragraph{Static and dynamic benchmarks.}
Mathematical reasoning benchmarks have progressed from grade-school word problems in GSM8K~\citep{cobbe2021training} through competition-level tasks in MATH~\citep{hendrycks2021measuring} to research-frontier difficulty in FrontierMath~\citep{glazer2024frontiermath}, with intermediate efforts spanning graduate-level questions~\citep{rein2024gpqa}, olympiad problems~\citep{he2024olympiadbench, gao2024omni, sun2025challenging, mahdavi2025leveraging}, hierarchical curricula~\citep{liu2024mathbench}, visual reasoning~\citep{lu2024mathvista}, and unified multi-task suites~\citep{mishra2022lila, frieder2023mathematical}.
All share a structural limitation: the problem pool is fixed at creation time, so leaderboards converge toward uninformative ceilings as models improve.
Dynamic alternatives partially address this. DynaBench~\citep{kiela2021dynabench} and Adversarial NLI~\citep{nie2020adversarial} have humans craft adversarial examples targeting current models; DyVal~\citep{zhu2023dyval} generates evaluation samples procedurally; perturbation-based methods~\citep{qian2024varbench} and transformation-based~\citep{zhou2025autocode} resist contamination by creating variants or seeding the existing problems; red-teaming with language models~\citep{perez2022red} replaces the human adversary with a model.
Live Benchmarks~\citep{white2024livebench, jain2024livecodebench, zheng2025livecodebench} and MathArena~\citep{balunovic2025matharena} address contamination by sourcing from recently released or continuously refreshed content but remain constrained by an external supply of problems.

\paragraph{Elo-based and preference-based evaluation.}
Chatbot Arena~\citep{chiang2024chatbot} introduced large-scale Elo ranking of LLMs through crowdsourced pairwise preference votes.
Closely related work studies LLM-as-a-judge directly~\citep{zheng2023judging}. Arena-Hard~\citep{li2024crowdsourced} and AlpacaEval~\citep{li2023alpacaeval} replace human judges with LLM judges for scalability, while JudgeBench~\citep{tan2024judgebench} reveals the limits of that substitution.
In game-playing AI, self-play Elo is the standard measure of strength: AlphaZero~\citep{silver2018general} rates agents by outcomes of games they play against each other, and adversarial self-play frameworks such as debate~\citep{irving2018ai, khan2024debating} and SPIN~\citep{chen2024self} extend the idea to language.
MathDuels adopts comparable pairwise-comparison machinery but grounds it in binary correctness on problems with verifiable answers rather than subjective preference, avoiding the reliability concerns of LLM-as-judge while retaining the comparative ranking structure.

\paragraph{Synthetic data.}
LLM-generated mathematical content has been used extensively for training: MetaMath~\citep{yu2024metamath} rewrites existing problems through forward and backward reasoning, WizardMath~\citep{luo2023wizardmath} applies Evol-Instruct to produce progressively harder items, MAmmoTH~\citep{yue2024mammoth} and OpenMathInstruct~\citep{toshniwal2024openmathinstruct} scale synthetic data to millions of examples, and Phi-4 shows the that carefully curated synthetic data can significantly improve reasoning ability~\citep{abdin2024phi}. MathDuels essentially employ a low-prior synthetic data generation process with specified domains and subdomains for the authoring stage.

\section{Conclusion}
\label{sec:conclusion}

We present MathDuels, a self-play benchmark that evaluates language models as both authors and solvers of mathematical problems, producing rankings whose difficulty co-evolves with participant strength rather than decaying as models improve past a fixed test set.
Experiments across a suite of frontier models show that authoring and solving capabilities are partially decoupled, and MathDuels rating can expose strength and capability gap in authoring math problems that solve-only benchmarks cannot capture. As LLMs start to help mathematicians attack open problems, and as their capabilities approach math frontiers which require both creativity and rigorous reasoning, we believe MathDuels will provide valuable insights on tracking and comparing progress across models.

Several directions for future work follow naturally from the MathDuels protocol. Scaling the problem budget per matchup would tighten confidence intervals on both rating estimates and ranking. Moving from final-answer checking to proof-based verification would let the benchmark reward rigorous reasoning rather than just correct outputs, opening the door to richer authoring strategies and finer-grained diagnostics. Perhaps most importantly, the self-play structure is domain-agnostic: the same author-solver framework can be applied to competitive programming, scientific question-answering, and other settings where generating hard, valid challenges is itself a meaningful test of understanding.
\newpage

\paragraph{Ethics statement}
This work studies fully-automated evaluation methodology for large language models and does not involve human subjects, personal data, or sensitive information.

\paragraph{Acknowledgments}
We thank the Google TPU Builders Program for providing the computational resources that supported this research.

\bibliography{colm2026_conference}
\bibliographystyle{colm2026_conference}

\newpage
\appendix

\section{Full Leaderboard}

\label{sec:appendix_rank_ranges}

Table~\ref{tab:rank_ranges_full} reports the full leaderboard with per-axis ratings, 95\% bootstrap CIs, composite rank ranges, and rank divergences at $k{=}30$ (19 models, 559 valid problems, 10{,}000 stratified bootstrap iterations).

\begin{table}[h]
\centering
\small
\begin{tabular}{@{}rlccccc@{}}
\toprule
\# & Model & Solve & Author & Composite & 95\% CI & Range \\
\midrule
1  & \texttt{Gemini-3.1-Pro-high}    & 2214 & 1624 & 1919 & [1856, 2000] & 1--2  \\
2  & \texttt{GPT-5.4-high}           & 2268 & 1473 & 1870 & [1798, 1975] & 1--3  \\
3  & \texttt{GPT-5.2-high}           & 2047 & 1427 & 1737 & [1675, 1807] & 2--6  \\
4  & \texttt{Claude-Opus-4.6-high}   & 2043 & 1307 & 1675 & [1622, 1740] & 3--8  \\
5  & \texttt{Gemini-3-Flash-high}    & 1944 & 1401 & 1673 & [1620, 1732] & 3--8  \\
6  & \texttt{GPT-5.4-low}            & 1958 & 1315 & 1637 & [1579, 1699] & 3--10 \\
7  & \texttt{Claude-Sonnet-4.6-high} & 1932 & 1254 & 1593 & [1533, 1658] & 4--12 \\
8  & \texttt{GPT-5.4-mini-high}      & 1922 & 1218 & 1570 & [1516, 1628] & 4--12 \\
9  & \texttt{Qwen-3.5-397B-A17B}     & 1972 & 1123 & 1548 & [1496, 1606] & 6--12 \\
10 & \texttt{Kimi-K2.5}              & 1925 & 1158 & 1542 & [1484, 1603] & 6--13 \\
11 & \texttt{Gemini-3.1-Pro-low}     & 1797 & 1208 & 1503 & [1452, 1555] & 7--14 \\
12 & \texttt{Grok-4.20-high}         & 1950 & 1020 & 1485 & [1443, 1534] & 7--14 \\
13 & \texttt{DeepSeek-V3.2}          & 1826 & 1072 & 1449 & [1408, 1492] & 10--15\\
14 & \texttt{GLM-5}                  & 1781 & 1095 & 1438 & [1397, 1482] & 11--15\\
15 & \texttt{MiniMax-M2.7}           & 1622 & 1122 & 1372 & [1324, 1426] & 13--18\\
16 & \texttt{Grok-4.1-fast-high}     & 1500 & 1169 & 1335 & [1292, 1386] & 15--19\\
17 & \texttt{GPT-5.4-mini-low}       & 1568 & 1091 & 1329 & [1287, 1374] & 15--19\\
18 & \texttt{Step-3.5-Flash}         & 1624 & 1032 & 1328 & [1290, 1370] & 15--19\\
19 & \texttt{Gemini-3-Flash-low}     & 1264 & 1226 & 1245 & [1178, 1303] & 16--19\\
\bottomrule
\end{tabular}
\caption{Full leaderboard at $k{=}30$ (19 models, 10{,}000 bootstrap iterations).
\textbf{Solve} and \textbf{Author}: Elo ratings on each axis.
\textbf{Composite}: weighted combination of both axes.
\textbf{95\% CI}: bootstrap confidence interval on the composite rating.
\textbf{Range}: best--worst composite rank consistent with bootstrapped CIs, assuming all other models' CIs take their most adversarial value.}
\label{tab:rank_ranges_full}
\end{table}

\section{Use of Large Language Models}
In this work, LLMs were used to assist with code development, figure generation, and language polishing.

\section{Problem Analysis}

\subsection{Problem Taxonomy}
\label{sec:appendix_taxonomy}

To ensure a broad coverage, each model's 30-problem generation budget is uniformly distributed across six broad mathematical domains via domain-specific meta-prompts (Table~\ref{tab:problem-taxonomy}). Aggregate solve rates reveal distinct variations in difficulty across these areas: Discrete Mathematics proved to be the most challenging domain for the models (82.5\% average accuracy), while Probability \& Statistics yielded the highest success rate (92.2\%).

\begin{table}[h]
\centering
\small
\begin{tabular}{p{0.23\linewidth} p{0.68\linewidth}}
\toprule
\textbf{Broad area} & \textbf{Subfields} \\
\midrule
Analysis & real analysis; measure and integration; functional analysis; PDEs; complex analysis \\[0.3em]

Algebra & linear algebra; abstract algebra (groups/rings/fields); representation theory; algebraic geometry; category theory \\[0.3em]

Geometry \& Topology & differential geometry; smooth manifolds; point-set topology; algebraic topology; homotopy theory \\[0.3em]

Discrete Mathematics & combinatorics; graph theory; logic and foundations; algorithms; complexity \\[0.3em]

Probability \& Statistics & probability theory; mathematical statistics; stochastic processes; stochastic calculus; Markov chains \\[0.3em]

Applied \& Computational Mathematics & differential equations; optimization; numerical analysis; dynamical systems; control theory \\
\bottomrule
\end{tabular}
\caption{Problem taxonomy. Each problem is assigned a broad area and a more specific subfield.}
\label{tab:problem-taxonomy}
\end{table}

\paragraph{Why discrete mathematics may be hardest in the arena.}

Discrete Mathematics has the lowest average solve rate in the arena. This may seem surprising, since discrete problems often look less technical or abstract than questions in areas such as algebraic geometry, algebraic topology, category theory, etc. We do not interpret this as an intrinsic ranking of mathematical fields. Rather, it likely reflects the kinds of problems the generation pipeline tends to produce in different domains. Many difficult discrete problems in our pool are short, finite, and easy to verify, but hard to solve because their relevant structure is not explicit in the statement. They force the solver to discover the right representation or counting framework before any standard method can begin. By contrast, more technical-looking problems in geometry, topology, or category theory are often harder to author correctly, since a novel question must respect a larger body of background structure. But once the intended machinery is recognized, the route to the answer can be comparatively direct.

\paragraph{Illustration.}
GPT-5.2-high provides a useful contrast. One of its more complex questions asks:

\emph{``Let
\[
F=\bigvee_{i=1}^4 S_i^2
\]
be the wedge of four copies of the $2$-sphere, form a simply connected CW-complex $X$ by attaching a $3$-cell along the class $2\alpha_1+2\alpha_3\in\pi_2(F,*)$, and let $f:(X,*)\to(X,*)$ be a based cellular homotopy equivalence inducing a specified homomorphism on $\pi_2(X,*)$. Define the mapping torus
\[
T_f=(X\times[0,1])/((x,1)\sim (f(x),0)).
\]
Determine the isomorphism type of $\pi_2(T_f,[*,0])$ as a finitely generated abelian group, written in invariant-factor decomposition.''}

This problem involves CW-complexes, attaching maps and induced maps on homotopy groups. It is mathematically delicate to author correctly, but once the solver identifies the intended machinery, the route is fairly structured. It has solve rate $17/19$. 

By contrast, the same model also asks:

\emph{``Consider a regular $12$-gon with vertices colored red, blue, and green so that exactly four vertices have each color, adjacent vertices are never the same color, and for each ordered pair of distinct colors $(X,Y)$, the number of indices $i$ such that vertex $i$ has color $X$ and vertex $i+1$ has color $Y$ is exactly $2$. Two colorings are identified up to dihedral symmetry. How many distinct colorings are there?''}

This problem is much easier to state and verify, but its difficulty is less visible. The solver must uncover the hidden combinatorial structure and then perform a nontrivial orbit count under symmetry. It therefore has a more adversarial flavor despite looking less technical. Its solve rate is only $11/19$.

\paragraph{Why Probability \& Statistics is easiest in the arena.}

Probability \& Statistics provides a useful contrast. In our pool, many problems in this domain reduce to familiar templates, especially conditioning arguments, symmetry, and standard finite probabilistic reasoning. Even when the statement is slightly dressed up, the underlying structure is often classical enough that once the right viewpoint is identified, the calculation is fairly direct.

\paragraph{Illustration.}
Claude-Opus-4.6-high asks:

\emph{``A standard 52-card deck is shuffled into a uniformly random order. Define event $A$ as the event that the first King in the deck appears in an earlier position than the second Queen (i.e., the Queen whose position is second-smallest among all four Queens), and define event $B$ as the event that the first King in the deck appears in an earlier position than the first Jack. Compute $P(A\mid B)$.''}

This is a clean and well-posed probability question, but its structure is quite classical. It has solve rate $19/19$.

\subsection{Authoring Styles: Textbook Recall vs. Originality}
\label{sec:appendix_authoring_styles}

\tcbset{
  problembox/.style={
    enhanced, breakable, colback=white, boxrule=0.8pt, arc=3pt,
    left=6pt, right=6pt, top=4pt, bottom=4pt,
    fonttitle=\small\bfseries, toptitle=3pt, bottomtitle=3pt
  }
}

Weak authors default to textbook exercises: standard objects, direct computations, memorized procedures. Strong authors impose structural conditions that require the solver to construct a solution path before any calculation is possible or combine multiple areas to create problems. 

\paragraph{Abstract Algebra}
\noindent Here the stronger author writes a problem whose difficulty lies in uncovering the right group-theoretic structure, whereas the weaker author asks for a straightforward execution of standard cyclotomic degree formulas.

\begin{tcolorbox}[problembox,
  colframe=blue!55!black,
  colbacktitle=blue!8!white,
  coltitle=blue!55!black,
  title={Strong author: \texttt{Gemini-3.1-Pro-high} \textnormal{\mdseries(Author rating 1624 $\cdot$ solve rate 8/19)}}]
\small
\textit{Find the total number of isomorphism classes of groups $G$ of order $2024$ such that every element of $G$ of odd order belongs to the commutator subgroup $G'$.}

\smallskip
\textbf{Answer:} $15$\quad The hypothesis forces the abelianization $G/G'$ to be a $2$-group, after which one has to classify the possible group structures compatible with that restriction.
\end{tcolorbox}

\smallskip

\begin{tcolorbox}[problembox,
  colframe=gray!55!black,
  colbacktitle=gray!12!white,
  coltitle=gray!55!black,
  title={Weak author: \texttt{Grok-4.20-high} \textnormal{\mdseries(Author rating 1020 $\cdot$ solve rate 19/19)}}]
\small
\textit{Let $\zeta$ be a primitive 91st root of unity. Compute $\left[\mathbb{Q}(\zeta^{13} + \zeta^{-13},\, \zeta^{7} + \zeta^{-7}) : \mathbb{Q}\right]$.}

\smallskip
\textbf{Answer:} $18$\quad Degrees of the real subfields are $\phi(7)/2 = 3$ and $\phi(13)/2 = 6$; coprimality gives the compositum degree by direct multiplication.
\end{tcolorbox}

\paragraph{Point-Set Topology} 
\noindent Here the stronger author turns point-set topology into a combinatorial problem, whereas the weaker author asks a definitional continuity check in a very standard pathological topology.

\begin{tcolorbox}[problembox,
  colframe=blue!55!black,
  colbacktitle=blue!8!white,
  coltitle=blue!55!black,
  title={Strong author: \texttt{Gemini-3.1-Pro-high} \textnormal{\mdseries(Author rating 1624 $\cdot$ solve rate 14/19)}}]
\small
\textit{Consider a set of 14 elements. Determine the maximum possible number of open sets a $T_0$ connected topology can have, given that its largest totally ordered subset in the specialization order contains exactly 3 elements.}

\smallskip
\textbf{Answer:} $2^{10} + 2^9 + 2^8 + 1$\quad Recast via the poset--topology correspondence as an extremal counting problem on a height-3 connected poset; no single theorem resolves it.
\end{tcolorbox}

\smallskip

\begin{tcolorbox}[problembox,
  colframe=gray!55!black,
  colbacktitle=gray!12!white,
  coltitle=gray!55!black,
  title={Weak author: \texttt{MiniMax-M2.7} \textnormal{\mdseries(Author rating 1122 $\cdot$ solve rate 19/19)}}]
\small
\textit{Let $X = \mathbb{R}$ with the lower limit topology (basis $\{[a,b) \mid a < b\}$) and $Y = \mathbb{R}$ with the Euclidean topology. Define $f : X \to Y$ by $f(x) = 0$ for $x \ge 0$ and $f(x) = 1$ for $x < 0$. Is $f$ continuous at $x_0 = 0$\,? Answer 1 if yes, 0 if no.}

\smallskip
\textbf{Answer:} $1$\quad A standard definition-check on a well-known pathological space; follows immediately from the basis criterion.
\end{tcolorbox}

\subsection{Failing on Self-Generated Problems}
\label{sec:appendix_author_blindness}

Models occasionally generate mathematically well-posed problems but compute an incorrect gold answer for their own creation. In these cases, the verification layer (Section~\ref{sec:judging}) overrides the author's solution if a correct mathematical consensus is found among the solvers.

For example, \texttt{DeepSeek-V3.2} authored the following functional analysis problem:

\begin{quote}
    \emph{``Consider the Hilbert space $L^2([0, 1])$ of square-integrable complex-valued functions on the interval $[0, 1]$ with respect to Lebesgue measure. The inner product is defined by $\langle f, g \rangle = \int_0^1 f(x)\overline{g(x)} dx$ for all $f, g \in L^2([0, 1])$, and the norm is $\|f\|_2 = \sqrt{\langle f, f \rangle}$. Define the Volterra operator $V : L^2([0, 1]) \to L^2([0, 1])$ by $V(f)(x) = \int_0^x f(t) dt$, for all $f \in L^2([0,1])$ and $x \in [0,1]$. Let $M$ be the subspace $M = \{ f \in L^2([0,1]) : \int_0^1 f(t) dt = 0 \}$. Compute the operator norm of $V$ restricted to $M$, that is, $\|V|_M\| = \sup \{ \|Vf\|_2 : f \in M, \|f\|_2 = 1 \}$.''}
\end{quote}

The authoring model submitted an incorrect gold answer of $\frac{2}{\sqrt{5}\pi}$. However, the problem formulation is valid and admits a standard reduction. Letting $u = Vf$, we have $u'(x) = f(x)$ almost everywhere with $u(0) = 0$. The constraint $f \in M$ enforces the right boundary condition $u(1) = \int_0^1 f(t)dt = 0$. Computing the operator norm therefore reduces to finding the supremum of $\|u\|_2$ over all $u \in H^1_0([0,1])$ subject to $\|u'\|_2 = 1$. By the Poincar\'e inequality for Dirichlet boundary conditions, this supremum is $1/\pi$.

Fifteen of the seventeen evaluating models correctly executed this reduction and output $1/\pi$, prompting the verifier to override the author's gold answer. This demonstrates that posing a mathematically coherent problem is a distinct capability from executing the standard deductions required to solve it.

A more elementary example comes from \texttt{Kimi-K2.5}:

\begin{quote}
\emph{``Determine the number of subsets $S \subseteq \{1,2,\ldots,12\}$ such that for all distinct $x,y\in S$, the absolute difference $|x-y|$ is not in $\{1,2,10,11\}$.''}
\end{quote}

The author's submitted answer was $199$, but the corrected gold answer is $98$ with pass rate 16/19.

\end{document}